\documentclass[runningheads]{llncs}
\usepackage{graphicx}
\usepackage{amsmath} 
\usepackage{amssymb}  
\usepackage{caption}
\usepackage{multirow}
\begin{document}
%
\title{Conv-MCD: A Plug-and-Play Multi-task Module for Medical Image Segmentation}
\titlerunning{Plug-and-Play Multi-task Module for Medical Image Segmentation }
%
\author{Balamurali Murugesan \thanks{Code and supplementary \url{https://github.com/Bala93/Multi-task-deep-network}} \inst{1,2} \orcidID{0000-0002-3002-5845} \and
Kaushik Sarveswaran \inst{2} \orcidID{0000-0002-1525-966X} \and
Sharath M Shankaranarayana \inst{3} \orcidID{0000-0001-6747-9050} \and 
Keerthi Ram \inst{2} \and 
Jayaraj Joseph \inst{2} \and 
Mohanasankar Sivaprakasam \inst{1,2}}
\authorrunning{B. Murugesan et al.}
%
\institute{Indian Institute of Technology Madras (IITM), India \and
Healthcare Technology Innovation Centre (HTIC), IITM, India \and Zasti, India \\
\email{balamurali@htic.iitm.ac.in}}
\maketitle              
%

\begin{abstract}

For the task of medical image segmentation, fully convolutional network (FCN) based architectures have been extensively used with various modifications. A rising trend in these architectures is to employ joint-learning of the target region with an auxiliary task, a method commonly known as multi-task learning. These approaches help impose smoothness and shape priors, which vanilla FCN approaches do not necessarily incorporate. In this paper, we propose a novel plug-and-play module, which we term as Conv-MCD, which exploits structural information in two ways - i) using the contour map and ii) using the distance map, both of which can be obtained from ground truth segmentation maps with no additional annotation costs. The key benefit of our module is the ease of its addition to any state-of-the-art architecture, resulting in a significant improvement in performance with a minimal increase in parameters. To substantiate the above claim, we conduct extensive experiments using 4 state-of-the-art architectures across various evaluation metrics, and report a significant increase in performance in relation to the base networks. In addition to the aforementioned experiments, we also perform ablative studies and visualization of feature maps to further elucidate our approach.
\vspace{-3mm}
\end{abstract}
\section{Introduction}
Segmentation is the process of extracting a particular region in an image and it is an essential task in medical image analysis. The extraction of these regions is challenging due to shape variations and fuzzy boundaries. Recently, deep learning networks like UNet \cite{unet} having encoder-decoder architecture with cross-entropy loss are used for medical image segmentation and have shown promising results. The two major drawbacks associated with these approaches are: 1) encoder-decoder networks suffer from structural information loss due to downsampling operations performed via max-pooling layers, and  2) cross-entropy loss is prone to foreground-background class imbalance problem. In many medical image applications, we will also be interested in multiple object instance segmentation. Also, the extracted region of interest will be used for diagnosis and surgery purposes, creating the need for outlier reduction. In this paper, we propose a module design which incorporates structural information as auxiliary tasks, inspired from the multi-task learning work \cite{caruana}. The module helps any state-of-the-art architecture handle structural information loss, reduce outliers, alleviate class imbalance and improve multi-instance object segmentation. We also show that learning a main task along with its related tasks will enable the model to generalize better on the original task because of its ability to learn common sub-features. 

Recently, multiple works have used multi-task learning to handle the structure information loss. Of these, DCAN \cite{dcan} and DMTN \cite{isbi_dcan} are of our interest. The commonality between DCAN and DMTN is their single encoder and two decoders architecture. The two decoders are used to learn multiple tasks at a time. In the case of DCAN, the mask is learned along with contour. Similarly, with DMTN, the mask is learned along with the distance map. The network DCAN provides additional information about the structure and shape through a parallel decoder to handle the information loss, but suffers from issues related to class imbalance similar to that of UNet. Both these class imbalance and structural information loss problems are overcome by the joint classification and regression based network DMTN. The mask predicted with this network also has reduced outliers compared to DCAN. But the network DMTN has difficulty in handling multi-instance object segmentation, wherein if one object is relatively small compared to the other object, it considers the small object as an outlier and removes it, which was not the case with UNet and DCAN. While the idea of learning contour and distance as a parallel task to the mask as done by DCAN and DMTN is appreciable, the main disadvantage stems from the architecture design. The task of extending the dual-decoder design to other popular architectures can be done in the following ways: 1) adding the auxiliary decoder part of DCAN or DMTN as a parallel decoder branch to the other base networks. 2) duplicating the already existing decoder block of base networks for some auxiliary task. Both these techniques increase the parameter count, memory usage and time consumption. 
\vspace{-4mm}
The key contributions of our paper are as follows:
\begin{itemize}
    \item We propose a novel module Conv-MCD (\textbf{M}ask prediction, \textbf{C}ontour extraction and \textbf{D}istance Map estimation). The module consists of three parallel convolutional filters to learn the three related tasks simultaneously. The proposed module handles class imbalance, reduces outliers, alleviates structural information loss and it works well with multi-instance object segmentation. 
    \item The proposed module can be added to any state-of-the-art base network with minimal effort. In this paper, we have added our module to some of them and compared it with the base networks. We observed that the networks with our module showed better results compared to the base networks. The same has been achieved with minimal increase in time, memory and number of parameters.
    \item To validate the performance improvement achieved from including our module, we conducted ablative studies, feature map visualization and validation error curve comparison. These studies showed that learning a major task in parallel with two related tasks reduces overfitting and enables the network to generalize well.  
\end{itemize}


\section{Methodology}
\begin{figure}
\centering
\includegraphics[width=0.6\linewidth]{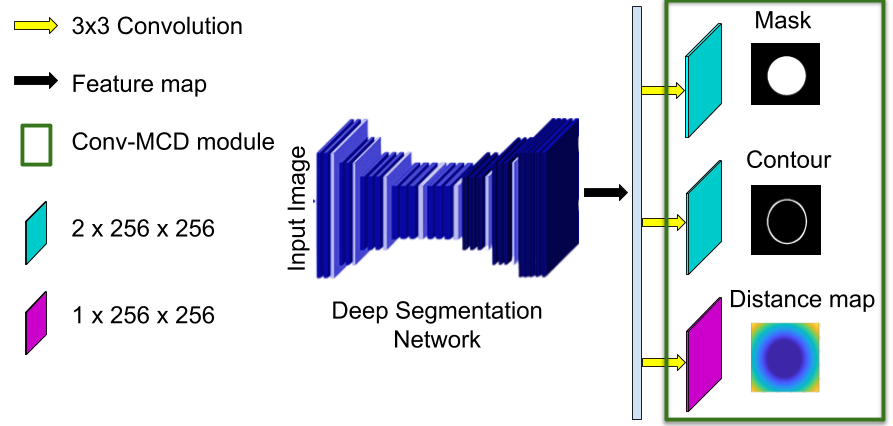}
\caption{Sample block diagram illustrating the proposed module Conv-MCD. Proposed module could be included at the end of a typical deep segmentation network.} \label{fig:module}
\end{figure}
\vspace{-10mm}
\subsection{Proposed module}
The proposed module Conv-MCD (Figure \ref{fig:module}) takes the feature maps from the deep learning networks as input and outputs mask, contour and distance map. This module helps the network to learn the multiple related tasks in parallel, enabling the network to generalize well. Mask prediction and contour extraction are classification tasks while the distance map estimation is a regression task. All these outputs are obtained by parallel convolution layers. The parameters of the filters are: kernel size is 3 $\times$ 3 with stride 1 and padding 1, the number of channels of the kernel is decided by the number of output channels of the feature maps. The number of filters for classification task is 2, which denotes the number of classes considered. Similarly, the number of filters for regression is 1. To show the effectiveness of the module with multi-task learning, we have considered only the binary classification problem. The same idea can be easily extended to multiple classes with appropriate change in module parameters. 

\subsection{Capturing Structural Information}\label{sect:structure_information}
We harness the structural information that is implicitly present in ground truth segmentation masks, which can be achieved in two ways:
\subsubsection{Contour extraction} 
For obtaining the contour map C, we first extract the boundaries of connected components based on the ground truth segmentation maps which are subsequently dilated using a disk filter of radius $5$. We empirically found that setting a radius of $5$ was optimal for an image of size $256\times256$. 
\subsubsection{Distance map estimation}
Using distance map alleviates the pixel-wise class imbalances which arise in the segmentation maps. We explore three kinds of distance maps and show that their choice is also an important factor in model performance. The distance maps D1 and D2 are obtained by applying euclidean distance transforms to mask and contour respectively. Distance map D3 is obtained by applying signed distance transform to the contour.
\subsection{Loss Function}
The loss function consists of three components - Negative Log Likelihood (NLL) loss for mask and contour, Mean Square Error (MSE) loss for the distance.  The total loss is given by 
\begin{equation}
\mathcal{L}_{total} = \lambda_{1} \mathcal{L}_{mask} + \lambda_{2} \mathcal{L}_{contour} + \lambda_{3} \mathcal{L}_{distance}    
\end{equation}
where $\lambda_{1},\lambda_{2},\lambda_{3}$ are scaling factors. 

The individual losses are formulated below:

\begin{equation}\label{eq:l_mask}
\mathcal{L}_{mask} = \sum_{\boldsymbol{x} \, \epsilon\, \Omega}log\, p_{mask}(\boldsymbol{x};l_{mask}(\boldsymbol{x}))
\end{equation}

\begin{equation}
\mathcal{L}_{contour} = \sum_{\boldsymbol{x}\, \epsilon\, \Omega}log\, p_{contour}(\boldsymbol{x};l_{contour}(\boldsymbol{x})) 
\end{equation}

\begin{equation}
\mathcal{L}_{distance} = \sum_{\boldsymbol{x}\, \epsilon\, \Omega} (\hat{D}(\boldsymbol{x}) - D(\boldsymbol{x}))^2 
\end{equation}

$\mathcal{L}_{mask}$, $\mathcal{L}_{contour}$ denotes the pixel-wise classification error. $\boldsymbol{x}$ is the pixel position in image space $\Omega$. $p_{mask}(\boldsymbol{x};l_{mask})$ and $p_{contour}(\boldsymbol{x};l_{contour})$ denotes the predicted probability for true label $l_{mask}$ and $l_{contour}$ after softmax activation function. $\mathcal{L}_{distance}$ denotes the pixel-wise mean square error. $\hat{D}(\boldsymbol{x})$ is the estimated distance map after sigmoid activation function while $D(\boldsymbol{x})$ is the ground-truth distance map.

\section{Experiments and Results}
\subsection{Dataset}
\textbf{Polyp segmentation}: We use Polyp segmentation dataset from MICCAI 2018 Gastrointestinal Image ANAlysis (GIANA) \cite{polyp_dataset}. We deemed this dataset to be ideal for our experiments since it exhibits the following characteristics: 1) large variations in shape 2) multi-instance object occurrences 3) foreground-background imbalance 4) difficulty in extracting the boundary of smooth, blobby objects. The dataset consists of 912 images with ground truth masks. The dataset is randomly split into 70\% for training and 30\% for testing. The images are center cropped and resized to 256 $\times$ 256.

\subsection{Implementation Details}
Models are trained for 150 epochs using Adam optimizer, with a learning rate of 1e-4 and batch size 4 in all the reported experiments, for consistent comparison. The train and validation loss plots can be found in supplementary material.
\subsection{Evaluation metrics}
The predicted and ground truth masks are evaluated using the following metrics: 1) \textbf{Segmentation evaluation}: Jaccard index and Dice similarity score are the most commonly used evaluation metrics for segmentation. 2) \textbf{Shape Similarity}: The shape similarity is measured by using the Hausdorff Distance (HD) between the shape of segmented object and that of the ground truth object. 3) \textbf{Segmentation around boundaries}: We evaluate the segmentation accuracy around the boundary with the method adopted in \cite{nips_trimap}. Specifically, we count the relative number of misclassified pixels within a narrow band (“trimap”) surrounding actual object boundaries, obtained from the accurate ground truth images. 4) \textbf{Boundary smoothness}: 
We extract the boundaries from the predicted mask and compare it with the ground truth boundaries using the maximum F-score (MF) as done in \cite{boundary_metric}.  



\subsection{Results and discussion}
\begin{figure}
\centering
\includegraphics[width=0.5\linewidth]{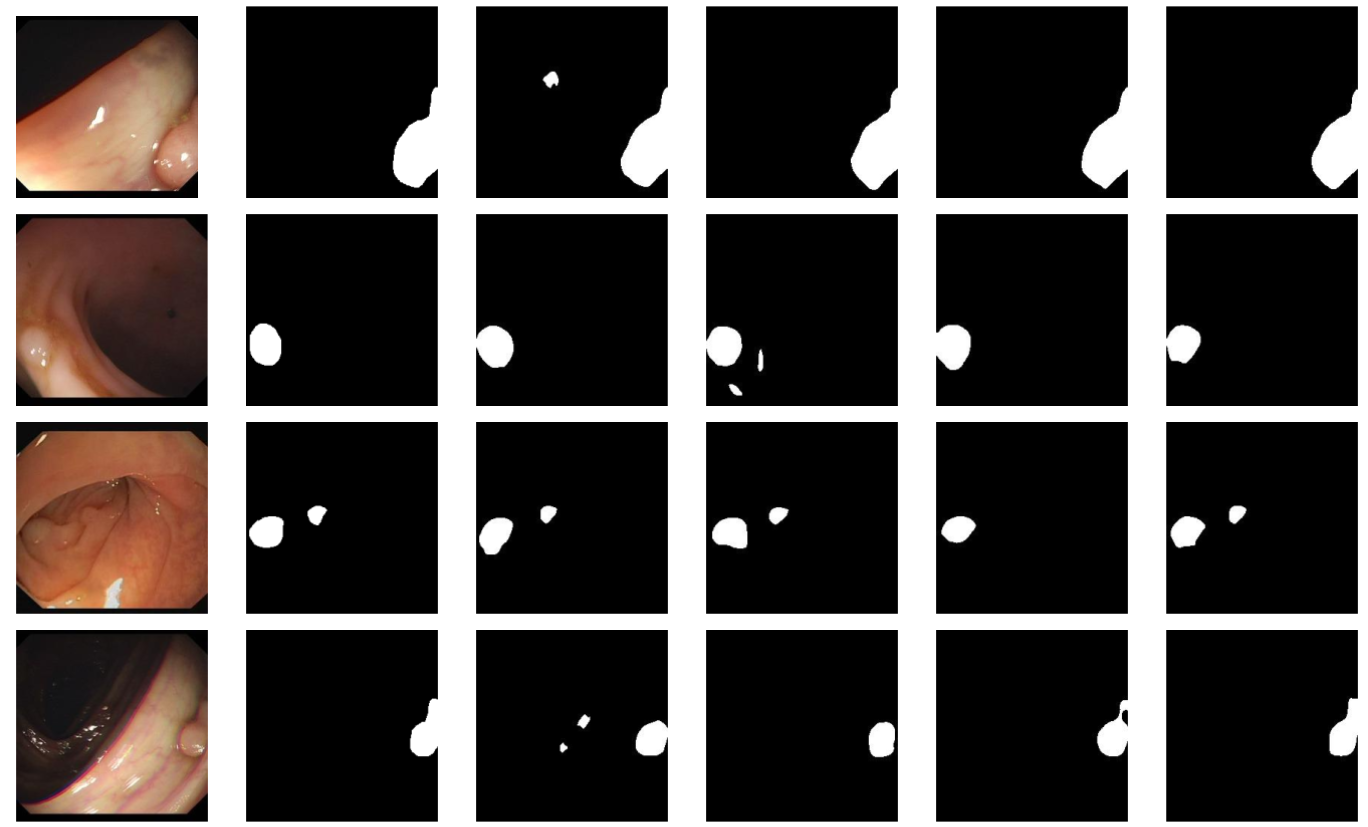}
\includegraphics[width=0.4\linewidth]{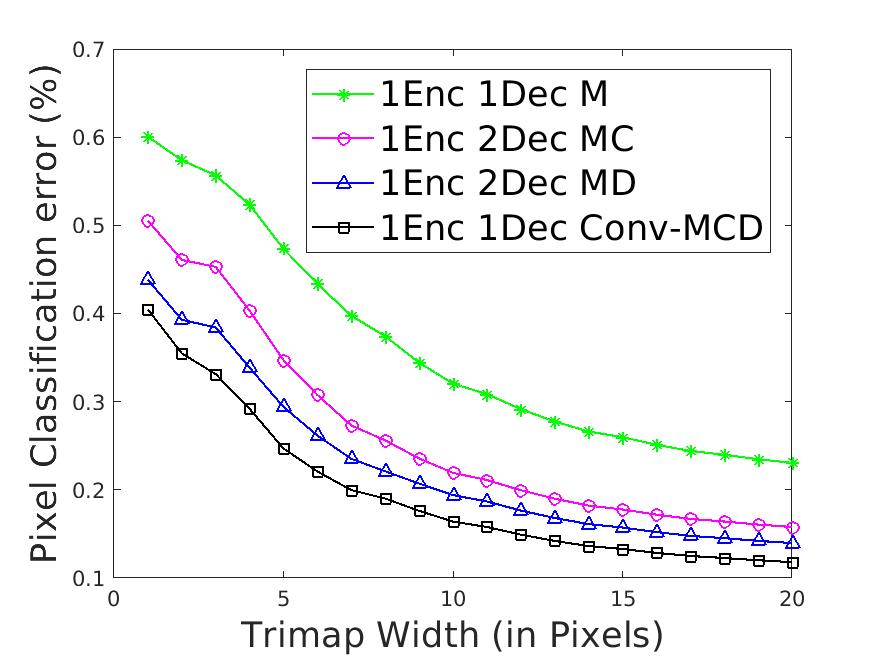}
\caption{Left (a): Four sample cases, from left to right: Image, Ground truth, 1Enc 1Dec M, 1Enc 2Dec MC, 1Enc 2Dec MD and 1Enc 1Dec Conv-MCD. Right (b): Pixel classification error vs trimap width for 1Enc 1Dec M, 1Enc 2Dec MC, 1Enc 2Dec MD and 1Enc 1Dec Conv-MCD.}
\label{fig:results1}
\end{figure}

Some notations used in this section are Encoder (Enc), Decoder (Dec), Mask (M), Contour (C), Distance (D) and our proposed module (Conv-MCD). The results of the network (1Enc 1Dec Conv-MCD) with the proposed module are compared with the following combinations of networks and loss functions: 1) A network (1Enc 1Dec M)  \cite{unet} with a single encoder and a decoder having NLL as loss function for mask estimation. 2) A network (1Enc 2Dec MC) \cite{dcan} with a single encoder and two decoders having NLL as loss function for both mask and contour estimation. 3) A network (1Enc 2Dec MD) \cite{isbi_dcan} with a single encoder and two decoders having NLL as loss function for mask and MSE as loss function for distance map estimation.
From Table \ref{table:metrics1}, it can be seen that the network 1Enc 1Dec Conv-MCD gives better segmentation, shape and boundary metrics compared to 1Enc 1Dec M, 1Enc 2Dec MC and 1Enc 2Dec MD with parameters and running time quite close to 1Enc 1Dec M. This brings the best of both worlds: better performance with lesser time and memory consumption. The graph in Figure \ref{fig:results1}b shows the network with our module is better than other networks at all trimap widths. From the figure, it can also be seen that 1Enc 2Dec MC and 1Enc 2Dec MD is better than 1Enc 1Dec M across all trimap widths. A similar observation can be found in the Table \ref{table:metrics1}, where 1Enc 1Dec MC and 1Enc 2Dec MD shows better metrics compared to 1Enc 1Dec M. This shows that contour and distance maps act as regularizers to the mask prediction. \footnote{To validate the generalisability of our proposed approach, we further test it against the baseline models on the ORIGA cup segmentation dataset. The quantitative results and observations on the same can be found in the supplementary material.} 
\begin{table}[h]
\centering
\caption{Comparison of 1Enc 1Dec Conv-MCD (ours) with \cite{unet},\cite{dcan},\cite{isbi_dcan}}
\begin{tabular}{|c|c|c|c|c|c|c|}
\hline
Architecture & Dice & Jaccard & HD & MF & Time (ms) & \# parameters \\ \hline
1Enc 1Dec M \cite{unet} & 0.8125 & 0.7323 & 24.13 & 0.6144 & 1.3131 & 7844256 \\ 
1Enc 2Dec MC \cite{dcan} & 0.8151 & 0.7391 & 22.74 & 0.616 & 1.8677 & 10978272 \\ 
1Enc 2Dec MD \cite{isbi_dcan} & 0.8283 & 0.7482 & 22.68 & 0.5681 & 1.8531 & 10977984 \\ 
1Enc 1Dec Conv-MC (Ours) & 0.8149 & 0.7389 & 22.86 & 0.6083 & 1.3384 & 7844832 \\ 
1Enc 1Dec Conv-MD (Ours) & 0.8286 & 0.7489 & 22.54 & 0.5844 & 1.3235 & 7844544 \\ 
1Enc 1Dec Conv-MCD (Ours) & \textbf{0.8426} & \textbf{0.7692} & \textbf{22.27} & \textbf{0.6552} & 1.3501  & 7845120 \\ \hline
\end{tabular}
\label{table:metrics1}
\end{table}

The qualitative comparison of the network 1Enc 1Dec Conv-MCD with other networks is displayed in Figure \ref{fig:results1}a. From first two rows of Figure \ref{fig:results1}a, it can be seen that the mask predictions by networks 1Enc 1Dec Conv-MCD and 1Enc 2Dec MD have smooth boundaries without outliers. The third row of the same figure depicts the case where 1Enc 2Dec MD fails, owing to its inability to segment multi-instance objects. The fourth row of the figure displays the case where 1Enc 1Dec Conv-MCD works better than the other networks. 

\begin{figure}
\centering
\includegraphics[width=0.4\linewidth]{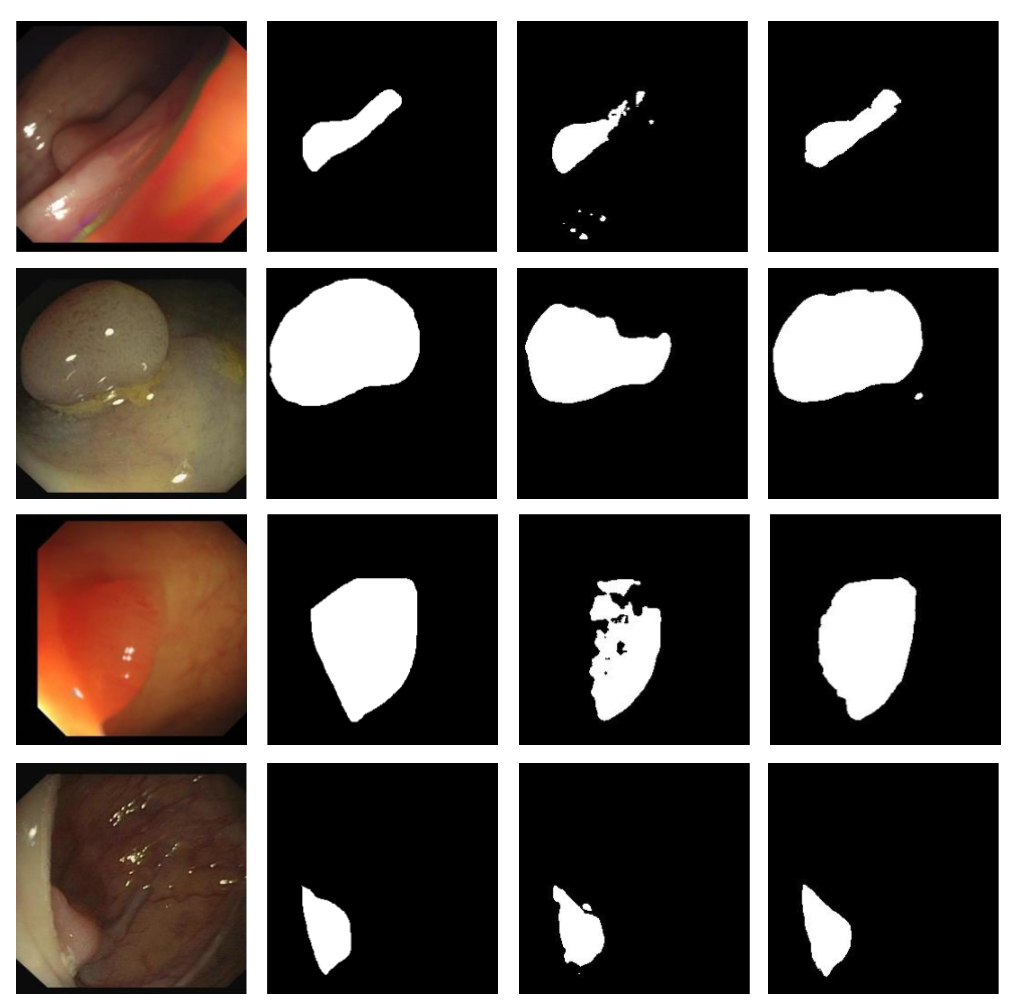}
\includegraphics[width=0.5\linewidth]{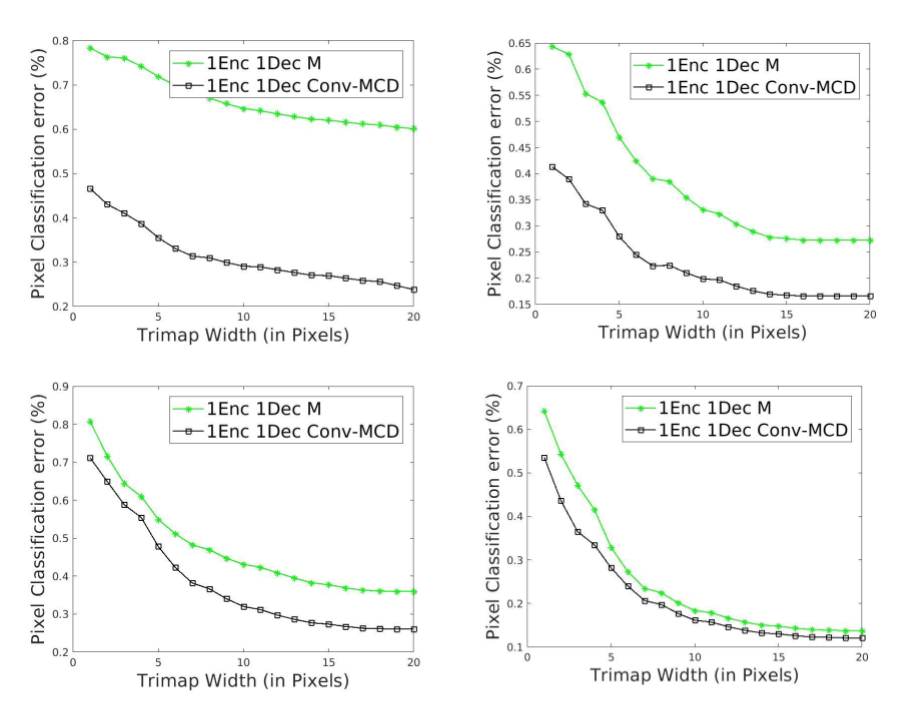}
\caption{Left (a): From row 1 to 4: SegNet, UNet, UNet16 and LinkNet34. In each row from left to right: Image, Ground Truth, network without our module Conv-MCD, network with our module Conv-MCD. Right (b): Row-wise order: Pixel classification error vs trimap width for SegNet, UNet, UNet16 and LinkNet34. In the graph, green line represents the base network and black line represents network with our module Conv-MCD.}

\label{fig:results2}
\end{figure}

Our proposed module is helpful in improving the performances of state-of-the-art segmentation networks. Adding our module to the segmentation networks is relatively simpler and is independent of the architecture design. Table \ref{table:metrics2} shows the performance of various state-of-the-art networks across different metrics, with and without our module. The networks used for comparison are UNet\cite{unet}, UNet16 (UNet with VGG16\cite{vgg} pre-trained encoder), SegNet\cite{segnet} and LinkNet34\cite{LinkNet}. It is evident that networks with our module gave improved results across all evaluation metrics. The qualitative comparison of these networks is shown in Figure \ref{fig:results2}a. From the figure, it can be observed that networks with our module were able to capture the shape, handle outliers better than networks without the module. In Figure \ref{fig:results2}b, it is shown that adding our module to the networks shows lower pixel classification error for different trimap widths compared to the base network.

\begin{table}[h]
\centering
\caption{Comparison of evaluation metrics for various state-of-the-art networks}
\begin{tabular}{|c|c|c|c|c|c|}
\hline
\multicolumn{1}{|l|}{Architecture} & Module type & Dice & Jaccard & HD & MF \\ \hline
\multirow{2}{*}{SegNet} & - & 0.6515 & 0.5497 & 41.31 & 0.4127 \\ 
 & Conv-MCD & \textbf{0.7194} & \textbf{0.6314} & \textbf{35.61} & \textbf{0.4923} \\ \hline
\multirow{2}{*}{UNet} & - & 0.8249 & 0.7468 & 23.27 & 0.6144 \\ 
 & Conv-MCD & \textbf{0.838} & \textbf{0.7652} & \textbf{21.05} & \textbf{0.6161} \\ \hline
\multirow{2}{*}{UNet16} & - & 0.8441 & 0.7676 & 15.24 & 0.7514 \\ 
 & Conv-MCD & \textbf{0.9124} & \textbf{0.8559} & \textbf{13.17} & \textbf{0.7753} \\ \hline
\multirow{2}{*}{LinkNet34} & - & 0.8835 & 0.8206 & 16.02 & 0.7358 \\ 
 & Conv-MCD & \textbf{0.8979} & \textbf{0.8383} & \textbf{14.28} & \textbf{0.7474} \\ \hline
\end{tabular}
\label{table:metrics2}
\end{table}
The selection of the type of distance map in the Conv-MCD module also affects model performance. We conducted experiments using 1Enc 1Dec Conv-MCD with different distance maps and found that D2 and D3 perform better than D1. Overall, 1Enc 1Dec Conv-MCD performs better than 1Enc 1Dec M for all types of distance maps. The results reported in this paper are from the best performing distance map. The comparison of distance map types can be found in the supplementary material.

To validate the effect of our module to the network performance, we conducted the following studies. In the first study, we plotted the validation loss values of the mask obtained from 1Enc 1Dec Conv-MCD and 1Enc 1Dec M against the number of epochs. The plot clearly showed that learning the additional tasks together did not allow the network to overfit, thus enabling the network to generalize better. As a second study, the feature maps extracted from the last layer of the network before the module are visualized. By visualization, it was found that while mask and contour had a direct representation, distance map could be represented by a linear combination of some feature maps. These feature representations demonstrate that a single decoder with our module will be sufficient instead of parallel decoders. The feature maps and the validation graphs can be found in supplementary material. The third one is an ablative study, where we compared our module with modules having a single auxiliary task, namely Conv-MC and Conv-MD. We observed that the network 1Enc 1Dec with Conv-MCD gives better results compared to Conv-MC and Conv-MD. This phenomenon can be attributed to increase in the number of related tasks \cite{caruana}. Quantitative results are available in Table \ref{table:metrics1}.

\section{Conclusion}
In this paper, we have proposed a novel plug-and-play module Conv-MCD. The module was specifically designed for medical images. It handles class imbalance, reduces outliers, alleviates structural information loss and works well with multi-instance object segmentation. Also, Conv-MCD can be added to any deep segmentation network, resulting in a significant improvement in performance with minimal increase in parameters.


\bibliographystyle{splncs04}
\bibliography{miccai}

\end{document}